\documentclass[10pt,twocolumn,letterpaper]{article}

\usepackage[pagenumbers]{cvpr} 
\usepackage{wrapfig}

\usepackage[dvipsnames]{xcolor}

\newcommand{\nerf}{NeRF}
\newcommand{\cliptonerf}{\texttt{clip2nerf}}
\newcommand{\nerftoclip}{\texttt{nerf2clip}}
\newcommand{\nftovec}{\texttt{nf2vec}}
\newcommand{\inrtovec}{\texttt{inr2vec}}
\newcommand{\shapenet}{Shape\-Net\-Ren\-der}

\usepackage{pifont}
\newcommand{\cmark}{\ding{51}}%
\newcommand{\xmark}{\ding{55}}%

\graphicspath{{./img/}}

\usepackage{bm}
\usepackage{float}
\usepackage{multirow}
\usepackage{physics}
\usepackage{comment}

\usepackage[accsupp]{axessibility}

\definecolor{cvprblue}{rgb}{0.21,0.49,0.74}
\usepackage[pagebackref,breaklinks,colorlinks,citecolor=cvprblue]{hyperref}

\def\paperID{1} 
\def\confName{CVPR}
\def\confYear{2024}

\title{Connecting NeRFs, Images, and Text}

\author{
Francesco Ballerini
\quad 
Pierluigi Zama Ramirez
\quad 
Roberto Mirabella
\\ \quad 
Samuele Salti
\quad 
Luigi Di Stefano 
\vspace{0.3em} \\
University of Bologna
\vspace{0.3em} \\
{\url{https://cvlab-unibo.github.io/clip2nerf}}
}

\begin{document}
\maketitle
\begin{abstract}
Neural Radiance Fields (NeRFs) have emerged as a standard framework for representing 3D scenes and objects, introducing a novel data type for information exchange and storage. Concurrently, significant progress has been made in multimodal representation learning for text and image data. This paper explores a novel research direction that aims to connect the NeRF modality with other modalities, similar to established methodologies for images and text. To this end, we propose a simple framework that exploits pre-trained models for NeRF representations alongside multimodal models for text and image processing. Our framework learns a bidirectional mapping between NeRF embeddings and those obtained from corresponding images and text. This mapping unlocks several novel and useful applications, including NeRF zero-shot classification and NeRF retrieval from images or text.
\end{abstract}
    
\section{Introduction}
\label{sec:intro}

In the Neural Radiance Fields (\nerf{}) framework \cite{nerf}, a neural network is trained to construct a volumetric representation of a 3D environment from images. Once a \nerf{} is trained, it enables the generation of novel views of that environment through ray tracing.
They have gained considerable popularity over recent years \cite{mittal2023neural}, emerging as a novel approach for 3D data representation.
Representing a scene with a single \nerf{} decouples the actual memory occupation from the spatial resolution and the number of observations. Indeed, we can encode a hypothetically infinite number of images at arbitrary resolution into a finite number of network weights.
This may potentially lead \nerf s to become a standard means of storing and exchanging 3D information, with entire databases of \nerf s residing on our hard drives in the future. Supporting this idea is the recent proliferation of various \nerf{} datasets \cite{de2023scannerf,ramirez2023deep,hu2023nerf}.

\begin{figure}[t]
    \centering
    \includegraphics[width=\columnwidth]{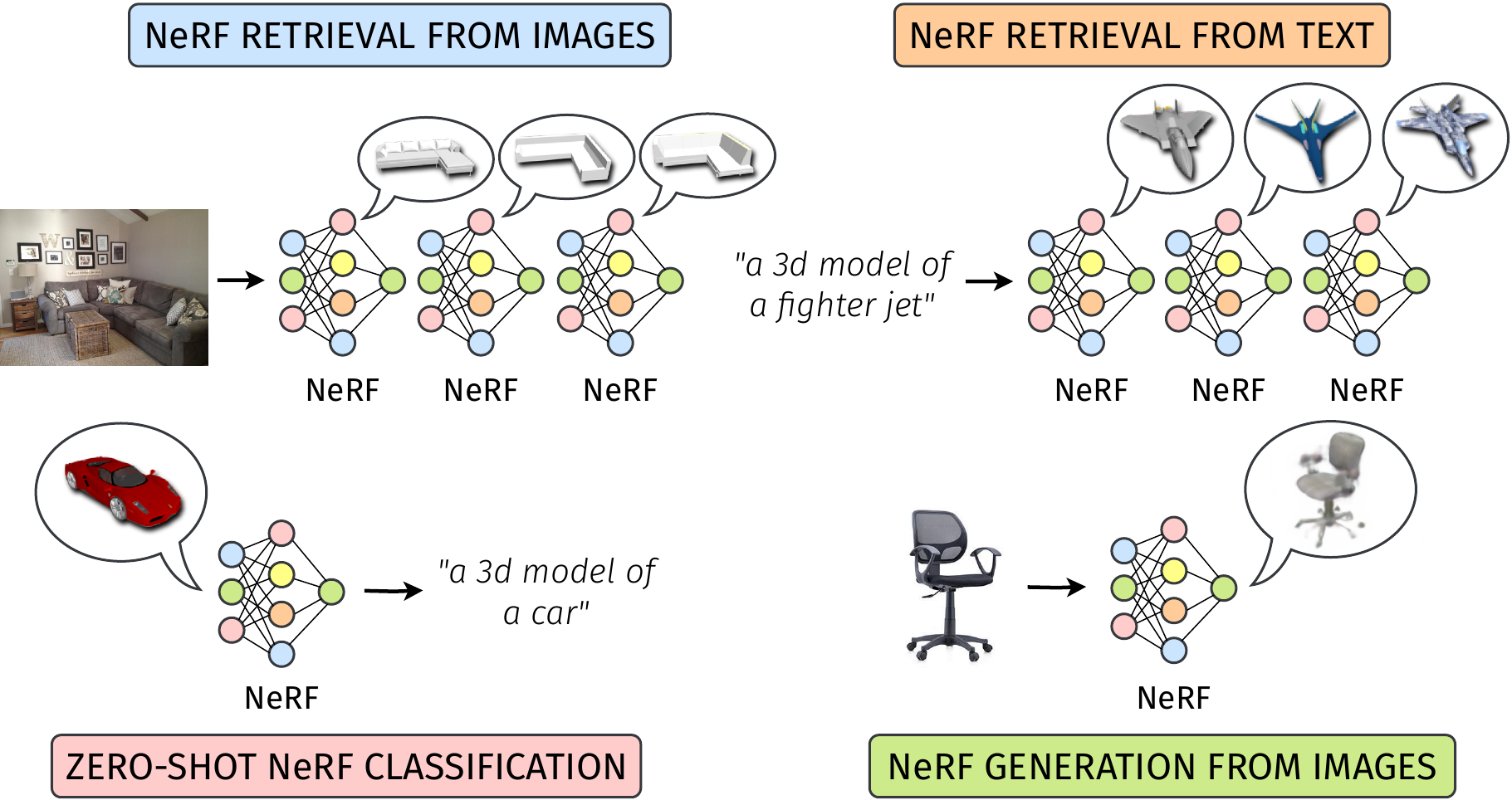}
    \caption{\textbf{Framework applications.} Examples of the possible tasks we can perform thanks to our framework that connects \nerf s, images, and text.}
    \label{fig:teaser}
\end{figure}

Concurrently with the development of \nerf s, there has been notable progress in the field of Vision-Language Models (VLMs) \cite{clip, girdhar2023imagebind}. These models capitalize on large paired databases of images and text to extract rich multimodal representations. 
By combining modalities, these representations obtain a better overall comprehension of images and text, allowing for better performance in existing visual and textual tasks. Moreover, they unlocked many novel applications such as zero-shot classification, where unseen instances are classified based on textual descriptions, image retrieval by querying with both text and image prompts, visual question answering, and many others. In the scenario in which \nerf s are an additional input modality, an intriguing research direction is to understand whether and how it is possible to connect \nerf s to other input modalities, as it was for images and text. By bridging \nerf s with diverse input modalities, we might unlock new opportunities for innovative applications.

Unlike images and text, which are well-studied input formats, \nerf s present unique challenges as they are neural networks, making them less straightforward to process by conventional frameworks.
One naive approach would involve rendering images of the object represented by a NeRF. However, this choice presents various challenges, including lengthy computation times, determining a viewpoint from which to render the object, or deciding on an appropriate rendering resolution.
Conversely, we would encounter none of these issues by processing the weights of the \nerf{}. The problem of how to process NeRF weights to enable downstream tasks has been the focus of the recently proposed framework \nftovec{} \cite{nf2vec}. This work learns to encode the information contained within the network weights into a compact embedding that retains sufficient knowledge to be used as input for downstream tasks.

Thanks to the availability of general purpose pre-trained VLMs such as CLIP \cite{clip} and pre-trained \nerf s encoders such as \nftovec{} \cite{nf2vec}, this paper casts the problem of connecting \nerf s, images, and text as learning mapping functions between the latent space of the \nftovec{} encoder and the latent space of CLIP.
In practice, we propose training two simple Multi-Layer Perceptrons (MLPs) to map a \nerf{} embedding into a CLIP embedding or vice versa. In this way, given an image or text input, we can discover the corresponding \nerf{} or vice versa. Notably, acquiring data for training such networks is straightforward, as we can directly leverage the renderings or ground-truth images of the \nerf s. Moreover, by exploiting a pre-trained model with multimodal text-image embeddings such as CLIP, we naturally learn the connection between \nerf{} and texts, avoiding the necessity of \nerf{}-text pairings.

Our framework unlocks many innovative and compelling applications, such as those depicted in \cref{fig:teaser}. For instance, it is possible to classify \nerf s in a zero-shot manner based solely on their weights, or given one or more images depicting an object, we can retrieve the most closely matching \nerf s. Alternatively, textual queries can be used to search for \nerf s stored in our databases. We can even generate entirely new \nerf s from either images or text.

Despite the simplicity of the architecture, we observe that our framework effectively performs tasks such as \nerf{} zero-shot classification on par with baselines operating on images obtained from \nerf s without requiring to render even a single pixel.
Moreover, leveraging recent text-to-image conditional generative approaches \cite{controlnet}, we propose an adaptation technique to apply our method effectively to real images even when trained solely on synthetic data.

Briefly, our contributions are:
\begin{itemize}
    \item We investigate for the first time the problem of connecting \nerf s with images and text.
    \item We propose the first framework to achieve this goal. Notably, this method is easy to train as it requires learning only two simple MLPs.
    \item Our idea unlocks many intriguing applications, such as zero-shot classification of \nerf s by solely processing their network weights and retrieving \nerf s from images or texts.
    \item We propose a technique to adapt our model to perform well on real images when trained solely on synthetic data.
\end{itemize}
\section{Related work}
\label{sec:related}

\paragraph{Vision-Language Models.} 
During the last few years, there has been a rapid advancement in visual-language modeling. The popularity of Vision Transformer (ViT) \cite{vit} has led to numerous studies that utilize ViT to simultaneously learn from vision-language data and achieve outstanding performance in downstream tasks \cite{su2019vl,lu2019vilbert,chen2020uniter,zhang2021vinvl,radford2021learning}. Researchers have proposed efficient pretraining tasks to enhance the alignment between visual and language modalities. Contrastive learning is one of the most prominent methods widely adopted in many studies \cite{clip,li2021align,he2020momentum,chen2020simple}. Among these methods, CLIP \cite{clip} is one of the most popular. Additionally, there are emerging works that explore unified frameworks to address vision-language tasks \cite{singh2022flava,lu2022unified,wang2022ofa,wang2023image,wang2021simvlm,wang2022git}. 
Recent works extend multimodal representation learning to other modalities such as audio and videos \cite{girdhar2023imagebind,wang2023one,wu2024nextgpt}. Our work employs CLIP to extract rich multimodal embeddings from images and text. 

\paragraph{Neural Radiance Fields.}
\nerf{} \cite{nerf} has emerged as a valuable tool for a variety of tasks, including view synthesis \cite{martin2021nerf}, generative media \cite{poole2022dreamfusion}, robotics \cite{yen2022nerf}, and computational photography \cite{mildenhall2022nerf}. Initially, the base \nerf{} model employed an MLP to translate spatial coordinates into color and density. Recent advancements substitute or enhance MLPs with voxel grid-like data structures \cite{Chen2022ECCV, sun2022direct, Plenoxels}. For instance, Instant NGP \cite{instant} utilizes a hierarchical arrangement of coarse and fine-grained grids stored using hashmaps. These structures facilitate the extraction of features, which are then processed by a compact MLP, resulting in significantly accelerated training processes. Our work employs \nerf s that follow the base formulation, \ie a single MLP extracting density and color information for each 3D coordinate.

\paragraph{Deep Learning on Neural Networks.}
Multiple recent studies have delved into using neural networks to process other neural networks. Early works in the field focused on forecasting network properties such as accuracy and hyper-parameters directly from their weights \cite{Unterthiner2020PredictingNN,urholt2021selfsupervised, knyazev2021parameter,jaeckle2021generating,Lu2020Neural}. Recent studies handle networks implicitly representing data (INRs or Neural Fields). These works perform vision tasks directly using network weights as the input or output data. Functa \cite{functa} learns priors across an entire dataset using a shared network and subsequently encodes each sample into a compact embedding employed for downstream discriminative and generative tasks. The following approaches focus on processing networks representing individual data, \eg a specific object or scene. The first framework doing it was \inrtovec{} \cite{deluigi2023inr2vec}. This approach encodes networks representing 3D shapes into compact embeddings, serving as input for subsequent tasks. \nftovec{} \cite{ramirez2023deep} extends \inrtovec{} to \nerf s, performing several tasks directly from \nerf s weights, such as classification, generation, or retrieval. \cite{cardace2024neural} learns how to process neural fields represented as a hybrid tri-plane representation. Another research direction \cite{navon2023equivariant, zhou2023neural, zhou2023permutation, zhou2024universal}, recognizing that MLPs exhibit weight space symmetries \cite{hecht1990algebraic}, proposes innovative architectures tailored for MLPs by leveraging network symmetries as an inductive bias. Other works \cite{kofinas2023graph, lim2024graph} exploit Graph Neural Networks to learn network representations. To improve the generalization of approaches processing neural networks, \cite{shamsian2024improved} explores various strategies for data augmentation directly in weight spaces. 
Our framework also processes neural networks representing individual objects. In particular, we employ \nftovec{} to extract rich embeddings from \nerf s.
\section{Connecting \nerf s and CLIP}
\label{sec:method}

Our work aims to learn the connection between image, text, and \nerf{} \cite{nerf} modalities. To achieve this goal, given rich multimodal representations extracted by Vision-Language Models such as CLIP \cite{clip} and compact embeddings extracted from \nerf{} weights by \nftovec{} \cite{nf2vec}, we learn how to map a \nftovec{} embedding into a plausible CLIP embedding and vice versa. In this section, we first report the relevant background knowledge: \nerf{}, \nftovec{}, and CLIP frameworks. Then, we describe our proposed framework depicted in \cref{fig:teaser}.

\subsection{Preliminaries}

\paragraph{\nerf{}.}
Given images of a scene or an object, \nerf{} \cite{nerf} allows for novel view synthesis from arbitrary vantage points. This is achieved by training a neural network, \ie an MLP, on a set of sparse images collected from different viewpoints.
We follow the base \nerf{} formulation \cite{nerf} in which a single MLP parameterizes the radiance field of the scene as a function of continuous 3D coordinates in space $\vb{x} = (x,y,z)$. Such a function produces a 4D output $RGB\sigma$, encoding the $RGB$ color and volume density $\sigma$ of each 3D point in the scene. 
$\sigma$ can be interpreted as the differential probability of a ray terminating at $\vb{x}$.
Given a \nerf{}, we can render an image from an arbitrary viewpoint with a desired resolution through volume rendering \cite{nerf}.
In our paper, \nerf s are considered a standard data format and the input to our framework. We assume that each \nerf{} encodes a specific object or scene. We wish to avoid sampling any information from the \nerf s, such as rendering views, as it would require vast computational overhead and pose many challenges, such as the choice of the rendering viewpoint. This work aims to extract all information solely by processing MLP weights of the \nerf{}.

\paragraph{\nftovec{}.}
\nftovec{} \cite{nf2vec} can extract compact embeddings from MLPs that parametrize \nerf s by processing only the network weights. These codes can then be processed using standard deep-learning pipelines to perform tasks such as classification or segmentation.
\nftovec{} is a representation learning framework that comprises an encoder and a decoder.
The encoder consists of a series of linear layers with batch normalization and ReLU non-linearity followed by a final max pooling. It processes each layer of the input MLP independently, obtaining one vector for each MLP layer. Then, the final max pooling compresses all the layer embeddings into one, obtaining the desired global latent vector representing the input MLP, \ie the input \nerf{}.
The decoder reproduces the original \nerf{} values given as input the embeddings produced by the encoder and a spatial coordinate $\mathbf{x}$.
Our paper utilizes the pre-trained \nftovec{} encoder to embed \nerf s, keeping it frozen.

\paragraph{CLIP.}
CLIP (Contrastive Language-Image Pre-training) \cite{clip} is a pioneering visual language representation model.
The CLIP architecture consists of an image and a text encoder such as ViT \cite{vit} and BERT \cite{devlin2018bert}, respectively. CLIP is trained using a contrastive learning objective on a large set of data, which encourages the model to assign similar embeddings to semantically related image-text pairs while maximizing the dissimilarity between embeddings of unrelated pairs. This procedure enforces a multimodal vision-language latent space in which images and corresponding textual prompts share the same embedding.
In our framework, we employ pre-trained and frozen CLIP encoders.

\subsection{Feature mapping networks}

\begin{figure}
    \centering
    \includegraphics[width=0.9\linewidth]{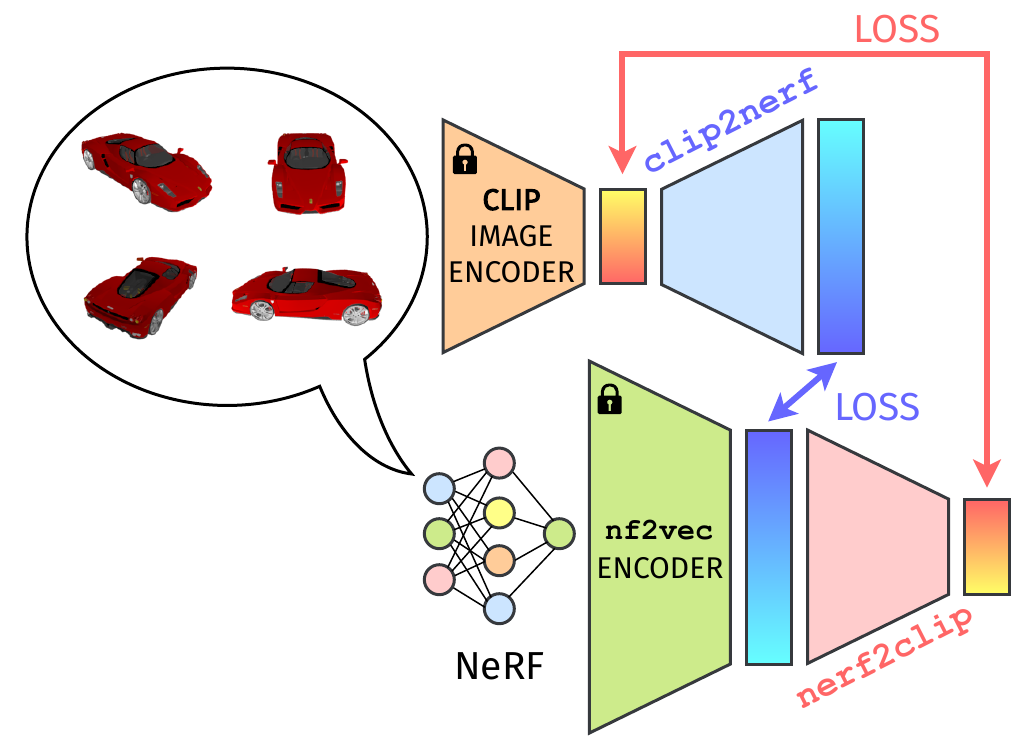}
   \caption{\textbf{Feature mapping network training.} \cliptonerf{} is a feature mapping network trained to map image embeddings of \nerf{} views to \nerf{} embeddings. Conversely, \nerftoclip{} computes the mapping in the opposite direction.}
   \label{fig:training}
\end{figure}

\paragraph{Architecture.}
To map CLIP embeddings to \nftovec{} embeddings, we use a simple MLP with GELU \cite{gelu} activation function and layer dimensions $512\rightarrow 768\rightarrow 1024$, where 512 and 1024 are the sizes of CLIP embeddings and \nftovec{} embeddings, respectively. We call this feature mapping network \cliptonerf{}. Similarly, we use another MLP, dubbed \nerftoclip{}, with GELU activation function and layer dimensions $1024\rightarrow 768\rightarrow 512$ to compute the mapping in the opposite direction.

\paragraph{Training.}
Given a \nerf{} and $n$ views of an object, we extract the \nftovec{} embedding $\vb{v}$ of the \nerf{} and $n$ CLIP embeddings $\vb{c}_i$ of its views.
For each view embedding $\vb{c}_i$, we train \cliptonerf{} to maximize the cosine similarity between its output 1024-dimensional vector $\hat{\vb{v}}_i$ and the embedding $\vb{v}$ of that \nerf{}. Formally, the \cliptonerf{} loss for an object is:
$$\mathcal{L}_\cliptonerf=\frac{1}{n}\sum_{i=1}^n\left({1-\frac{\hat{\vb{v}}_i \cdot \vb{v}}{\|\hat{\vb{v}}_i\|\|\vb{v}\|}}\right)$$
Instead, we train \nerftoclip{} to map a NeRF embedding $\vb{v}$ to the \emph{mean} embedding of the $n$ views, ${\vb{c} = \frac{1}{n}\sum_{i=1}^n\vb{c}_i}$, as learning to map a $\vb{v}$ to every $\vb{c}_i$ would create a one-to-many correspondence, \ie not a function. Specifically, we maximize the cosine similarity between the \nerftoclip{} 512-dimensional output $\hat{\vb{c}}$ and $\vb{c}$. Formally:
$$\mathcal{L}_\nerftoclip={1-\frac{\hat{\vb{c}} \cdot \vb{c}}{\|\hat{\vb{c}}\|\|\vb{c}\|}}$$
During training, the $n$ views can be the ground-truth images used to train the \nerf{} or those rendered from it. 

\section{Experimental Settings}
\label{sec:exp-settings}

\paragraph{\nerf{} framework and dataset.} We use the \nerf{} \cite{nerf} dataset of \nftovec{} \cite{nf2vec}. The \nerf{} architecture consists of an MLP with ReLU activation function and 4 hidden layers with 256 features each. It applies a frequency encoding \cite{nerf} $\text{enc}(\vb{x})$ to each input 3D coordinate $\vb{x}$. Each \nerf{} is trained on $N=36$ views of a shape of the \shapenet{} dataset \cite{shapenet_render}. The dataset consists of a \nerf{} for each \shapenet{} shape, for a total of 38653 \nerf{}s, which we split into training (30946), validation (3847), and test (3860) sets. For each \nerf{}, we have access to the 36 synthetic images used for training and their depth maps.

\paragraph{Metrics.} We evaluate our retrieval experiments (\cref{sec:retrieval}) with the recall@$k$ metric, \ie the percentage of queries $q$ such that at least one among the first $k$ neighbors of $q$ share the same label as $q$. The top-$k$ nearest neighbors of $q$ are those with the highest cosine similarity with $q$, sorted from closest to furthest. We call them 1-NN, 2-NN, \dots, $k$-NN. Our classification experiments use the standard multi-class accuracy (\cref{sec:zero-shot}).

\paragraph{Training details.} Our feature mapping networks \cliptonerf{} and \nerftoclip{} are trained for 150 and 100 epochs, with learning rates $10^{-5}$ and $10^{-3}$, respectively. Both are trained with the AdamW optimizer \cite{adamw}, one-cycle learning rate scheduler \cite{onecycle}, weight decay $10^{-2}$, and batch size 64. We perform all our experiments on a single NVIDIA RTX 3090 GPU.

\section{Zero-shot \nerf{} classification}
\label{sec:zero-shot}

\begin{figure}
    \centering
    \includegraphics[width=0.7\linewidth]{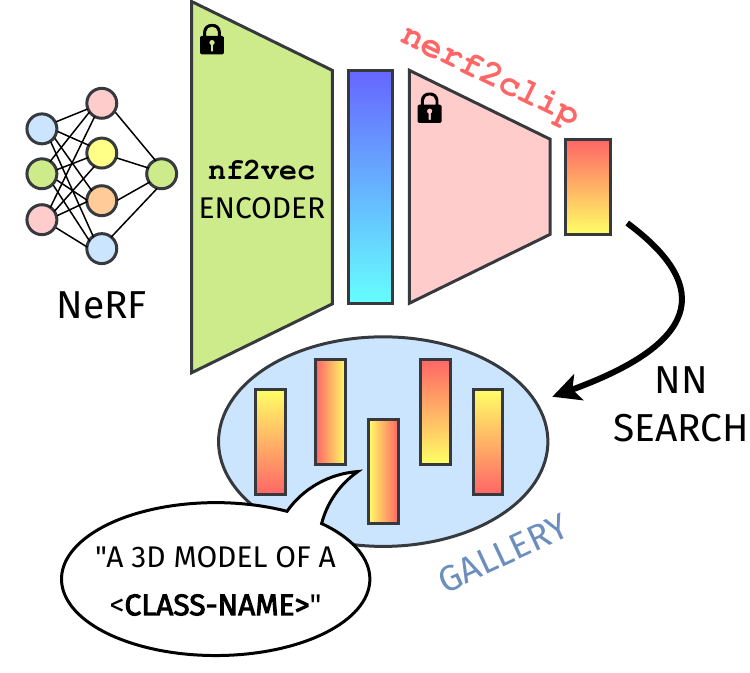}
    \caption{\textbf{Zero-shot \nerf{} classification method overview.}}
    \label{fig:zero-shot-method}
\end{figure}

\begin{figure}
    \centering
        \resizebox{\columnwidth}{!}{
        \begin{tabular}{lcccc}
            \toprule
            Method & Supervision & Rendering & Accuracy (\%) $\uparrow$ & Time (ms) $\downarrow$\\
            \midrule
            CLIP \cite{clip} 1 view & \multirow{6}{*}{\xmark} & \multirow{6}{*}{\cmark} & 73.6 & 13 \\
            CLIP \cite{clip} 2 views & & & 77.7 & 25 \\
            CLIP \cite{clip} 4 views & & & 80.5 & 49 \\
            CLIP \cite{clip} 8 views & & & 81.7 & 97 \\
            CLIP \cite{clip} 16 views & & & 82.4 & 193\\
            CLIP \cite{clip} $N$ views & & & 82.4 & 433 \\
            \cmidrule(lr){1-5}
            \nerftoclip{} rendered (ours) & \multirow{2}{*}{\xmark} & \multirow{2}{*}{\xmark} & 83.0 & \multirow{2}{*}{\textbf{2}} \\
            \nerftoclip{} GT (ours) & & & \textbf{84.0} & \\
            \cmidrule(lr){1-5}
            \nftovec{} \cite{nf2vec} (oracle)& \cmark & \xmark & 87.3 & 1 \\
            \bottomrule
        \end{tabular}}
        \captionof{table}{\textbf{Zero-shot \nerf{} classification results.}}
        \label{tab:zero-shot}
    \textcolor{white}{\rule{2cm}{0.4cm}}
        \includegraphics[width=\columnwidth]{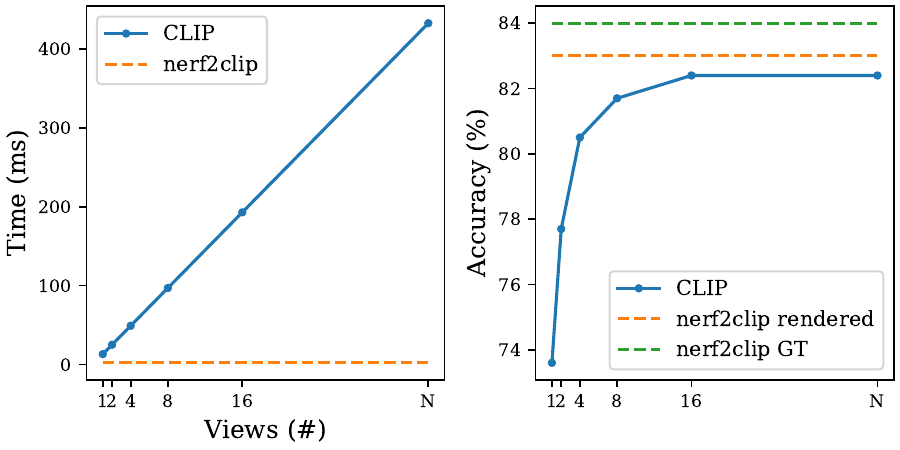}
        \captionof{figure}{\textbf{Zero-shot \nerf{} classification results (plots).} Classification time and accuracy as a function of the number of views used for the query.}
        \label{fig:zero-shot-plot}
\end{figure}

To perform zero-shot \nerf{} \cite{nerf} classification, we build a gallery of CLIP \cite{clip} embeddings of sentences of form ``A 3D model of \texttt{<class-name>}'', where \texttt{<class-name>}'' denotes a class label of \shapenet{} \cite{shapenet_render}. In other words, the gallery contains one CLIP embedding for each \shapenet{} class. We then take a test-set \nerf{}, encode it with the \nftovec{} \cite{nf2vec} encoder, process the results with \nerftoclip{}, and use the output embedding to query the gallery. Finally, the predicted label corresponds to the text of the 1-NN. This procedure is illustrated in \cref{fig:zero-shot-method}. As in our retrieval experiments (\cref{sec:retrieval}), the 1-NN maximizes the cosine similarity with the query. \cref{tab:zero-shot} shows classification results for two variants of \nerftoclip{}, one trained with ground-truth views (``\nerftoclip{} GT'' row) and the other with views rendered from the corresponding \nerf{} (``\nerftoclip{} rendered'' row). As an ablation, \cref{tab:zero-shot-abl} shows the effect of training \nerftoclip{} with an increasing number of views, \ie training \nerftoclip{} to map a \nerf{} embedding to the mean CLIP embedding of $n$ \nerf{} views, where $n=1,2\dots N$. The model with the highest accuracy in \cref{tab:zero-shot-abl} is the one reported in \cref{tab:zero-shot}.

As a baseline to compare \nerftoclip{} with, we query our gallery of CLIP class-label embeddings with the mean of the CLIP embeddings of $n$ random rendered views of a test-set \nerf{}, where $n=1,2\dots N$. 
We do not use ground-truth views as queries, as we assume they are only available at training time. 
The results are shown in \cref{tab:zero-shot} (``CLIP'' rows). \nerftoclip{}, in both its versions, achieves higher accuracy than the baseline, which plateaus at 16 views. For comparison, we also report the classification accuracy of \nftovec{}. It is important to note, however, that \nftovec{} performs classification in a supervised manner, and therefore its accuracy should be regarded as an upper bound to our zero-shot classification results. 

Furthermore, \cref{tab:zero-shot} compares the time required to perform zero-shot classification with \nerftoclip{} vs.\ the CLIP baseline. For the baselines, we report the sum of view rendering time, CLIP inference time, and NN search time. The rendering and the CLIP inference times scale linearly with the number of views. For \nerftoclip{}, we report the sum of \nftovec{} encoder inference time, \nerftoclip{} inference time, and NN search time. Our method is even faster than the CLIP baseline using only 1 view.

The results of \cref{tab:zero-shot} are also depicted in \cref{fig:zero-shot-plot}, where we highlight the fact that, while time and accuracy are a function of the number of views used to query the gallery for the CLIP baseline, they are a constant for \nerftoclip{} at inference time, as the query is the output of the trained model, thus requiring no rendering nor CLIP inference. Furthermore, as already pointed out, \nerftoclip{} achieves the highest accuracy and the fastest inference time.

\begin{table}
  \centering
  \resizebox{0.7\columnwidth}{!}{
  \begin{tabular}{lcc}
    \toprule
    Method & Views & Accuracy (\%) $\uparrow$\\
    \midrule
    \multirow{6}{*}{\nerftoclip{} rendered} & 1 & 80.6 \\
    & 2 & 81.4 \\
    & 4 & 82.4 \\
    & 8 & 82.3 \\
    & 16 & \textbf{83.0} \\
    & $N$ & 82.7 \\
    \cmidrule(lr){1-3}
    \multirow{6}{*}{\nerftoclip{} GT} & 1 & 82.9 \\
    & 2 & 83.2 \\
    & 4 & \textbf{84.0} \\
    & 8 & 83.8 \\
    & 16 & 83.6 \\
    & $N$ & 82.7 \\
    \bottomrule
  \end{tabular}}
  \caption{\textbf{\nerftoclip{} training ablation.} Effect of the number of views used for the feature mapping network training on the classification accuracy.}
  \label{tab:zero-shot-abl}
\end{table}

\section{\nerf{} Retrieval}
\label{sec:retrieval}

\begin{figure}
    \centering
    \includegraphics[width=\columnwidth]{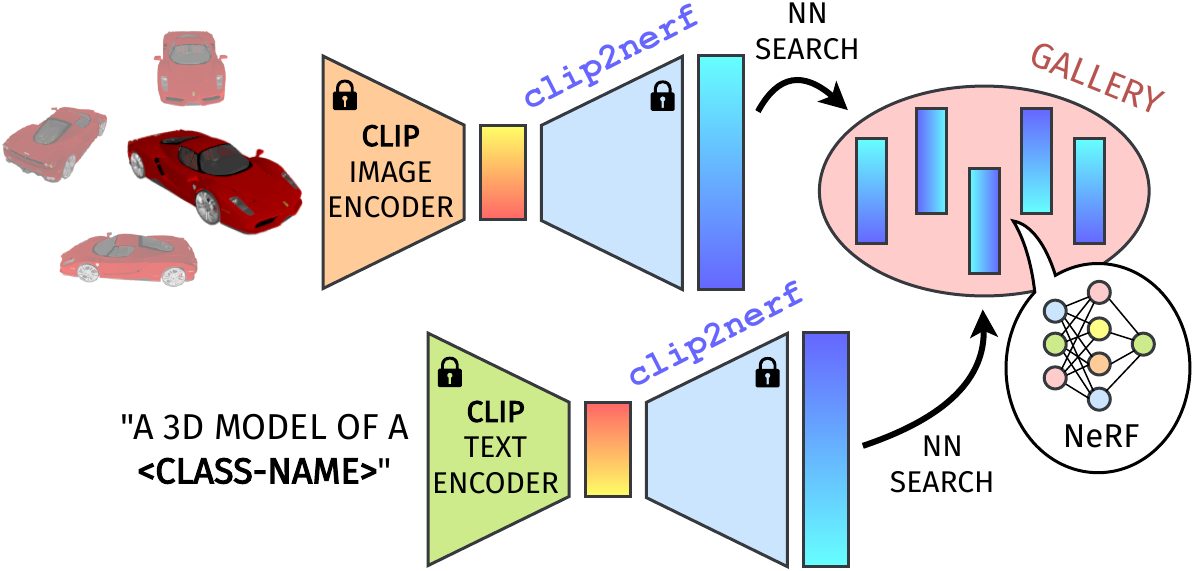}
    \caption{\textbf{\nerf{} retrieval method overview.} \nerf{} retrieval from images (top) and text (bottom).}
      \label{fig:retrieval_method}
\end{figure}

In the \nerf{} \cite{nerf} retrieval application, we aim to identify \nerf s in a database that closely matches a given textual or image query. To achieve this, we first construct offline a gallery of \nerf{} embeddings obtained by \nftovec{} \cite{nf2vec}. Then, we employ the image or text CLIP \cite{clip} encoders to generate the corresponding embedding during retrieval. This is processed by \cliptonerf{}, yielding the predicted \nerf{} embedding. Finally, we employ the NN search to locate the closest embedding within the gallery. This procedure is illustrated in \cref{fig:retrieval_method}. 

\subsection{\nerf{} retrieval from images}
\label{sec:retr-img}

The experiments reported here address the scenario in which a user takes one or more pictures of an object (\eg a product in a store) and uses them to query a database of \nerf{} objects (\eg the online store catalog). Thus, we always employ real or synthetic images as queries, \ie no rendered images from \nerf s.
In the experiments of this section, we build the gallery with the \nerf{} embeddings obtained by \nftovec{} on the test set of \shapenet{} \cite{shapenet_render}. We exclude the queried \nerf{} from the gallery during retrieval.

\paragraph{Single-view query.}
In \cref{tab:retr-single}, we report the results using a single image as query (a random GT view of the test set for each object). We use the same random images across different table rows to compare methods fairly.
We report the results of our \cliptonerf{} method, trained using either the ground-truth images (\cliptonerf{} GT) or the rendered images from the \nerf s (\cliptonerf{} rendered). In this way, we simulate the two possible scenarios in which the images used to train the \nerf s are available or not at training time.
\begin{table}[t]
  \centering
  \resizebox{\columnwidth}{!}{
  \begin{tabular}{lccccc}
    \toprule
    & \multicolumn{3}{c}{Recall (\%) $\uparrow$} & & \\
    \cmidrule(lr){2-4}
    Method & @1 & @5 & @10 & Time (ms) $\downarrow$ & Memory (MB) $\downarrow$ \\
    \midrule
    CLIP \cite{clip} GT mean & 81.4 & 93.9 & \textbf{96.3} & \multirow{2}{*}{24} & \multirow{2}{*}{8} \\
    CLIP \cite{clip} rendered mean & 81.2 & 92.9 & 96.1 & & \\
    \cmidrule(lr){1-6}
    CLIP \cite{clip} GT all & 83.6 & 93.0 & 95.1 & \multirow{2}{*}{331} & \multirow{2}{*}{271} \\
    CLIP \cite{clip} rendered all & 81.6 & 91.7 & 95.1 & & \\
    \cmidrule(lr){1-6}
    \cliptonerf{} GT (ours) & \textbf{86.1} & \textbf{94.0} & 96.0 & \multirow{2}{*}{25} & \multirow{2}{*}{15} \\
    \cliptonerf{} rendered (ours) & 84.5 & 93.3 & 95.4 & & \\
    \bottomrule
  \end{tabular}}
  \caption{\textbf{\nerf{} retrieval from images (single-view query results).}}
  \label{tab:retr-single}
\end{table}
\begin{figure}[t]
    \centering
    \includegraphics[width=\linewidth]{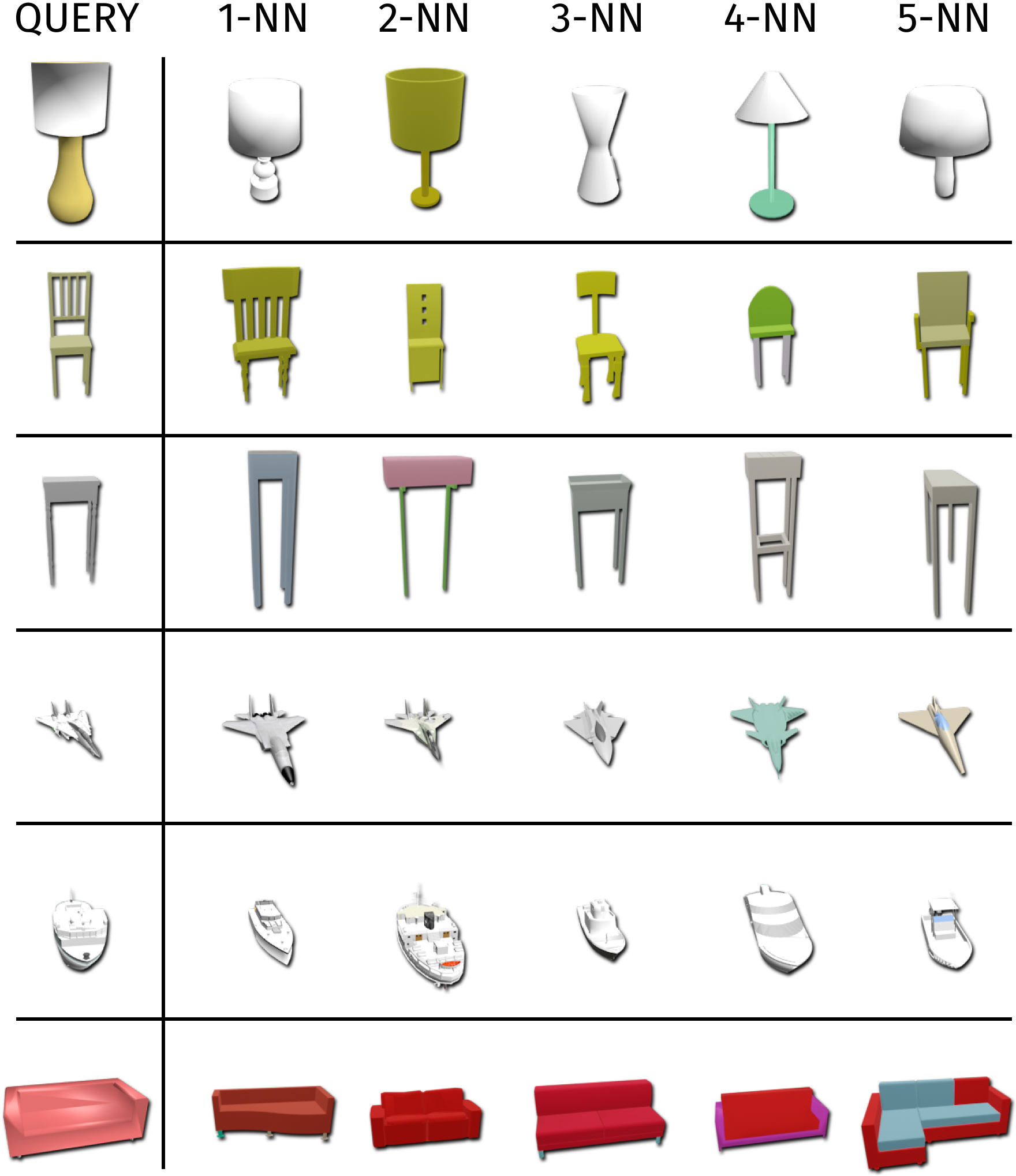}
    \caption{\textbf{Qualitative results of \nerf{} retrieval from images on \shapenet{} {\normalfont\cite{shapenet_render}}.} For each \nerf{}, we visualize one view rendered from a vantage point.}
      \label{fig:retrieval_from_images}
\end{figure}
Moreover, we report the performance of four plausible baseline strategies that exploit different galleries built using CLIP embeddings from images. ``CLIP GT mean'': each gallery element is the average of $N=36$ CLIP embeddings obtained from $N$ object views. ``CLIP rendered mean'': the same as the previous one, yet we employ $N$ rendered images from \nerf{} from the same $N$ viewpoints. ``CLIP GT all'': for each object in the test set, we store $N$ CLIP embeddings in the gallery, one for each ground-truth view. ``CLIP rendered all'': the same as the previous one, yet it employs the rendered views from \nerf{} instead of ground-truth images.
As shown in \cref{tab:retr-single}, our model exhibits superior performance in terms of recall@1 compared to the baselines while maintaining comparable results in the other metrics. 
%
Finally, we show some retrieval examples in \cref{fig:retrieval_from_images}. We render a single reference view to visualize \nerf s. As we can see, the retrieved \nerf s belong to the same object class and resemble the input image in color and shape.

\begin{figure}
  \centering
   \includegraphics[width=\columnwidth]{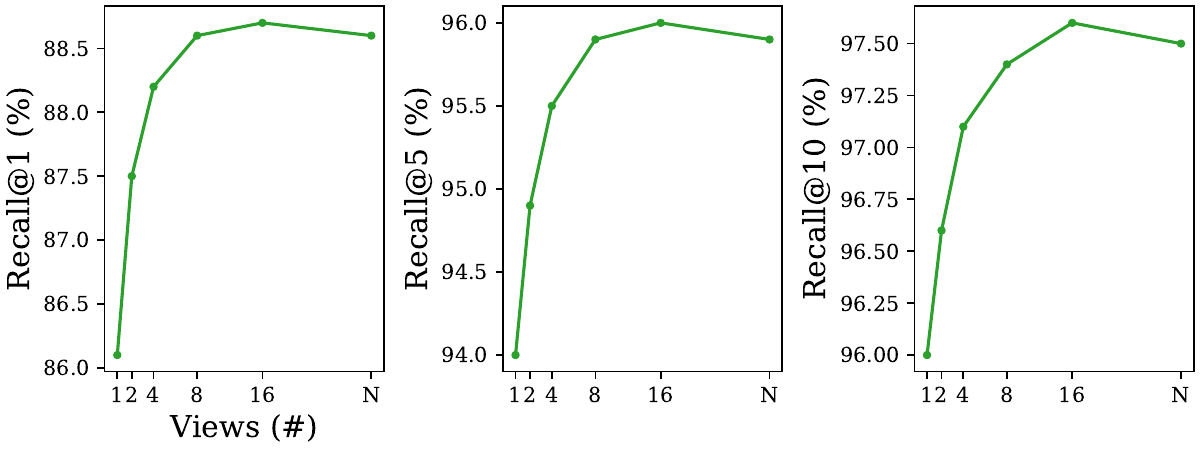}
   \caption{\textbf{\nerf{} retrieval from images (multi-view query results).} Performance of our \cliptonerf{} feature mapping network as a function of the number of views used for the query.}
   \label{fig:retr-multi}
\end{figure}
\begin{figure}
    \centering
    \includegraphics[width=0.9\linewidth]{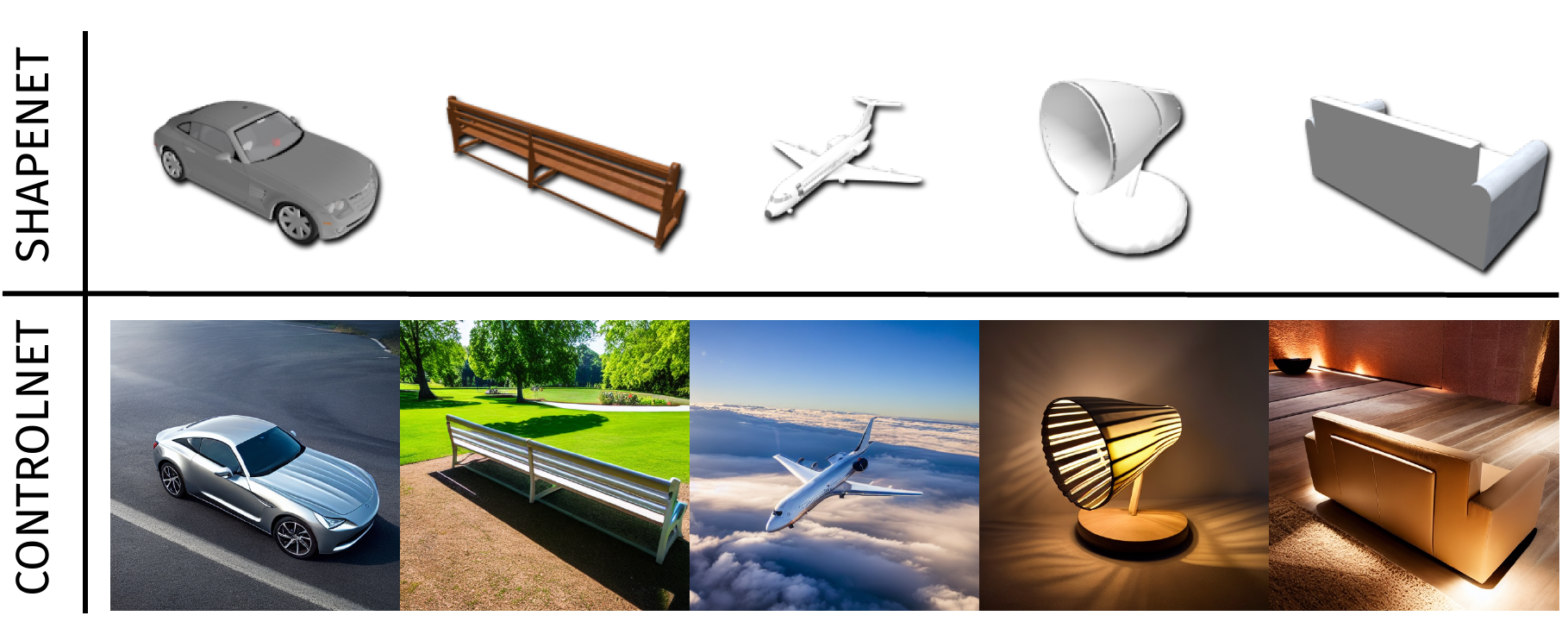}
    \caption{\textbf{ControlNet generated views.} ShapeNetRender \cite{shapenet_render} views vs.\ their counterparts generated by ControlNet \cite{controlnet}.}
    \label{fig:controlnet}
\end{figure}

\paragraph{Multi-view query.}
We focus here on the case where a user can acquire multiple pictures of the same object. In the plots of \cref{fig:retr-multi}, we show the retrieval recall@1, recall@5, and recall@10 results when varying the number of query images used for each object.
The gallery is the same as in the single-view scenario, \ie a \nerf{} embedding for each object. 
The query views are selected randomly among the ground-truth images in \shapenet{}. We randomly choose only the additional views when increasing the number of queries. For instance, the experiment with 8 views includes the images used in the 4 views results.
To retrieve the \nerf{}, we pass the $n$ multi-view queries to the CLIP image encoder, obtaining $n$ embeddings. Each is processed by \cliptonerf{} (the model named \cliptonerf{} GT in \cref{tab:retr-single}), and the resulting \nerf{} embeddings are averaged to get a reference embedding for that object. Then, we perform the NN search within the gallery.
Interestingly, when the number of query images used for retrieval increases, the results improve significantly until 8 views, at which point performance plateaus. Thus, we can conclude that the information provided by the additional views can be valuable for retrieval.

\begin{figure}
    \centering
    \includegraphics[width=\linewidth]{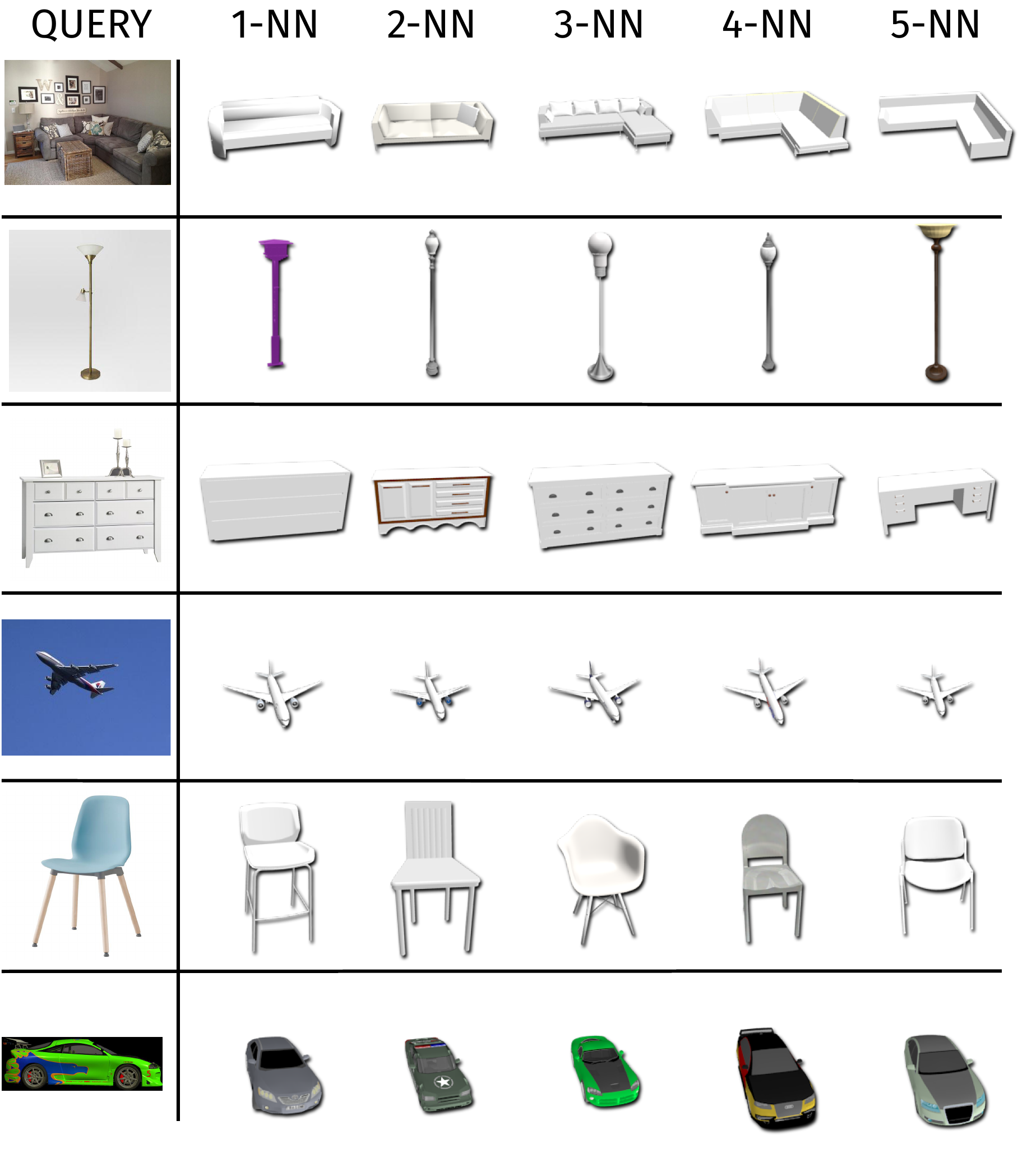}
    \label{fig:dom-adapt}
    \caption{\textbf{Qualitative results of \nerf{} retrieval from real images.} Queries are real images from DomainNet \cite{domainnet}. For each \nerf{}, we visualize one view rendered from a vantage point.}
\end{figure}

\begin{table}[t]
    \centering
    \begin{tabular}{lcccc}
        \toprule
        & \multicolumn{3}{c}{Recall (\%) $\uparrow$} \\
        \cmidrule(lr){2-4}
        Method & @1 & @5 & @10 \\
        \midrule
        CLIP \cite{clip} GT & 75.5 & \textbf{90.4} & 93.7 \\
        CLIP \cite{clip} rendered & 73.9 & 89.8 & \textbf{93.8} \\
        \cliptonerf{} GT (ours) & 67.9 & 80.7 & 85.6 \\
        \cliptonerf{} GT syn2real (ours) & \textbf{79.9} & 87.4 & 90.1 \\
        \bottomrule
    \end{tabular}
    \caption{\textbf{\nerf{} retrieval from real images (adaptation results).} Gallery of \nerf s from \shapenet{} \cite{shapenet_render}. Queries from DomainNet \cite{domainnet}.}
    \label{tab:dom-adapt}
\end{table}

\paragraph{Adaptation to real images.}
In the previous retrieval experiments, we employed solely synthetic query images. However, in a practical scenario, we would use real images acquired in the wild. Thus, we evaluated \cliptonerf{} using the real split of the DomainNet \cite{domainnet} dataset, reporting results in \cref{tab:dom-adapt}. The gallery consists of the \nerf{} embeddings from \shapenet{}. 
We note a performance drop compared to the case of testing on synthetic images, probably due to the domain-shift problem.
Thus, we propose an adaptation protocol based on the recent diffusion-based conditional generative approach, ControlNet \cite{controlnet}. In particular, we generate a new \emph{synthetic to real} datasets with ControlNet, using the synthetic object depth map as input to the generative network (see \cref{fig:controlnet}). These generated images are added to the synthetic ones to train \cliptonerf{}. The augmented dataset contains 7 synthetic random views and 7 images generated by ControlNet for each object.
By training on this dataset, we learned a feature mapping that can be applied effectively to real images. We report the performance of this network in the last row of \cref{tab:dom-adapt}, and we can observe a remarkable performance improvement w.r.t.\ the network trained without augmented data. Moreover, our method performs comparably to the baseline using the CLIP galleries obtained from images (rows 1 and 2 vs.\ 4).
Finally, in \cref{fig:dom-adapt}, we show retrieval results using in-the-wild real queries. Remarkably, the retrieved \nerf{} resembles the geometry of the input image with high fidelity.

\begin{figure}
    \centering
    \includegraphics[width=\linewidth]{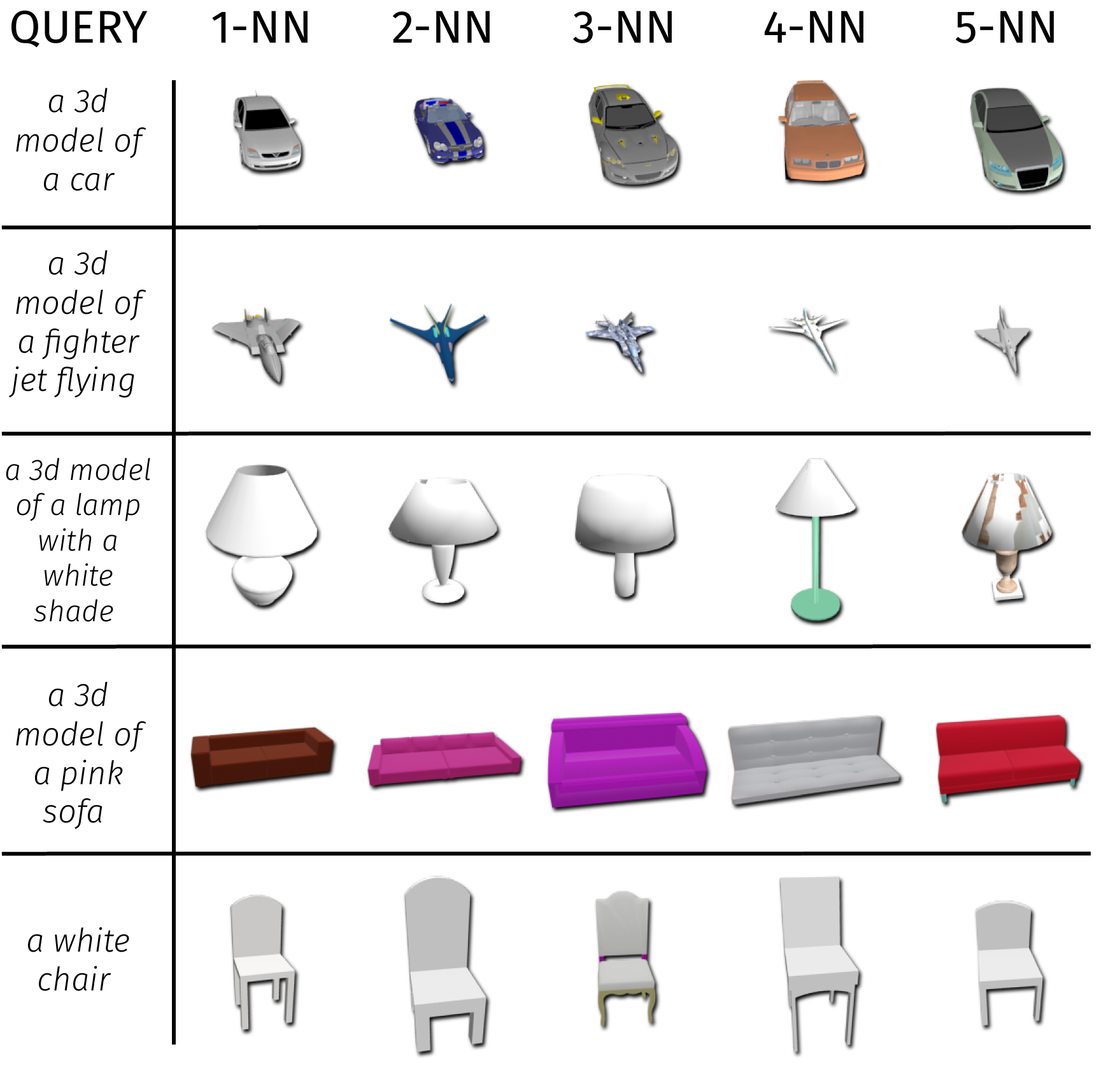}
    \caption{\textbf{Qualitative results of \nerf{} retrieval from text on \shapenet{} {\normalfont\cite{shapenet_render}}.} For each \nerf{}, we visualize one view rendered from a vantage point.}
    \label{fig:retrieval_from_text}
\end{figure}

\begin{table}
  \centering
  \resizebox{\columnwidth}{!}{
  \begin{tabular}{lccc}
    \toprule
    & \multicolumn{3}{c}{Recall (\%) $\uparrow$} \\
    \cmidrule(lr){2-4}
    Method & @1 & @5 & @10 \\
    \midrule
    CLIP GT & 80.0 & 89.6 & 92.8 \\
    CLIP rendered & 79.6 & 89.5 & 92.0 \\
    \cliptonerf{} GT (ours) & 63.2 & 75.2 & 79.0 \\
    \cliptonerf{} GT multimodal (ours) & \textbf{85.6} & \textbf{91.7} & \textbf{93.3} \\
    \bottomrule
  \end{tabular}}
  \caption{\textbf{\nerf{} retrieval from text results.}}
  \label{tab:retr-text}
\end{table}

\begin{figure}[t]
    \centering
    \includegraphics[width=0.9\columnwidth]{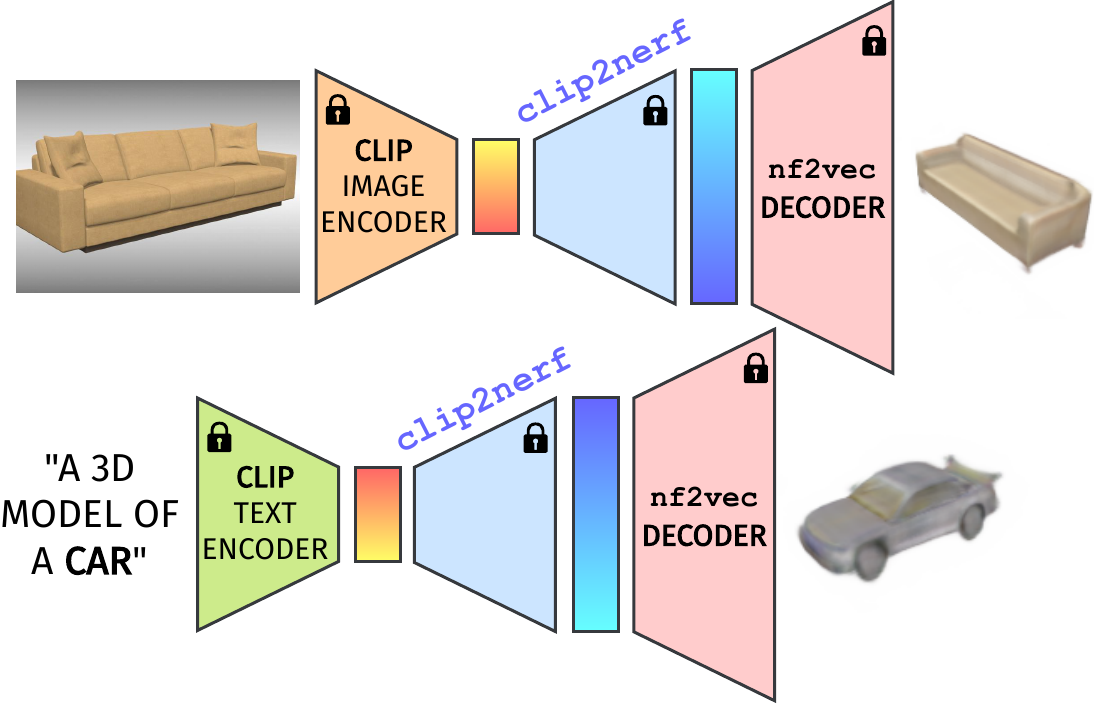}
    \caption{\textbf{\nerf{} generation method overview.} \nerf{} generation from images (top) and text (bottom).}
    \label{fig:gen-method}
    \vspace{-1mm}
\end{figure}

\subsection{\nerf{} retrieval from text}
\label{sec:retr-text}

We experiment here with the retrieval of \nerf s from text. In this scenario, given a text prompt, we want to find the corresponding \nerf{} in our database. 

We employ the same gallery of the single-view scenario of \nerf{} retrieval from images. To obtain a reasonable query text for the input images, we use the BLIP-2 \cite{blip2} captioner.
We report results on \shapenet{} in \cref{tab:retr-text}. Our \cliptonerf{} obtains lower performance than the baselines using the CLIP galleries. We relate this to the feature mapping function, which was never trained on the CLIP text embeddings. For this reason, we train a variant of our method using as input to \cliptonerf{} either the CLIP embeddings obtained from an image or the CLIP embedding of an automatically generated caption with BLIP-2. As shown in \cref{tab:retr-text}, this multimodal training paradigm can even surpass the baselines by a moderate margin (row 1 and 2 vs.\ row 4).

Finally, we also visualize some qualitative results in \cref{fig:retrieval_from_text}. We note that we can retrieve \nerf s of the class described in the text, which contains details presented in the textual prompt, \eg we correctly retrieve \nerf s of a jet fighter in the second row.

\section{\nerf{} generation}
\label{sec:generation}

\begin{figure}[t]
    \vspace{-1mm}
    \centering
    \includegraphics[width=0.9\columnwidth]{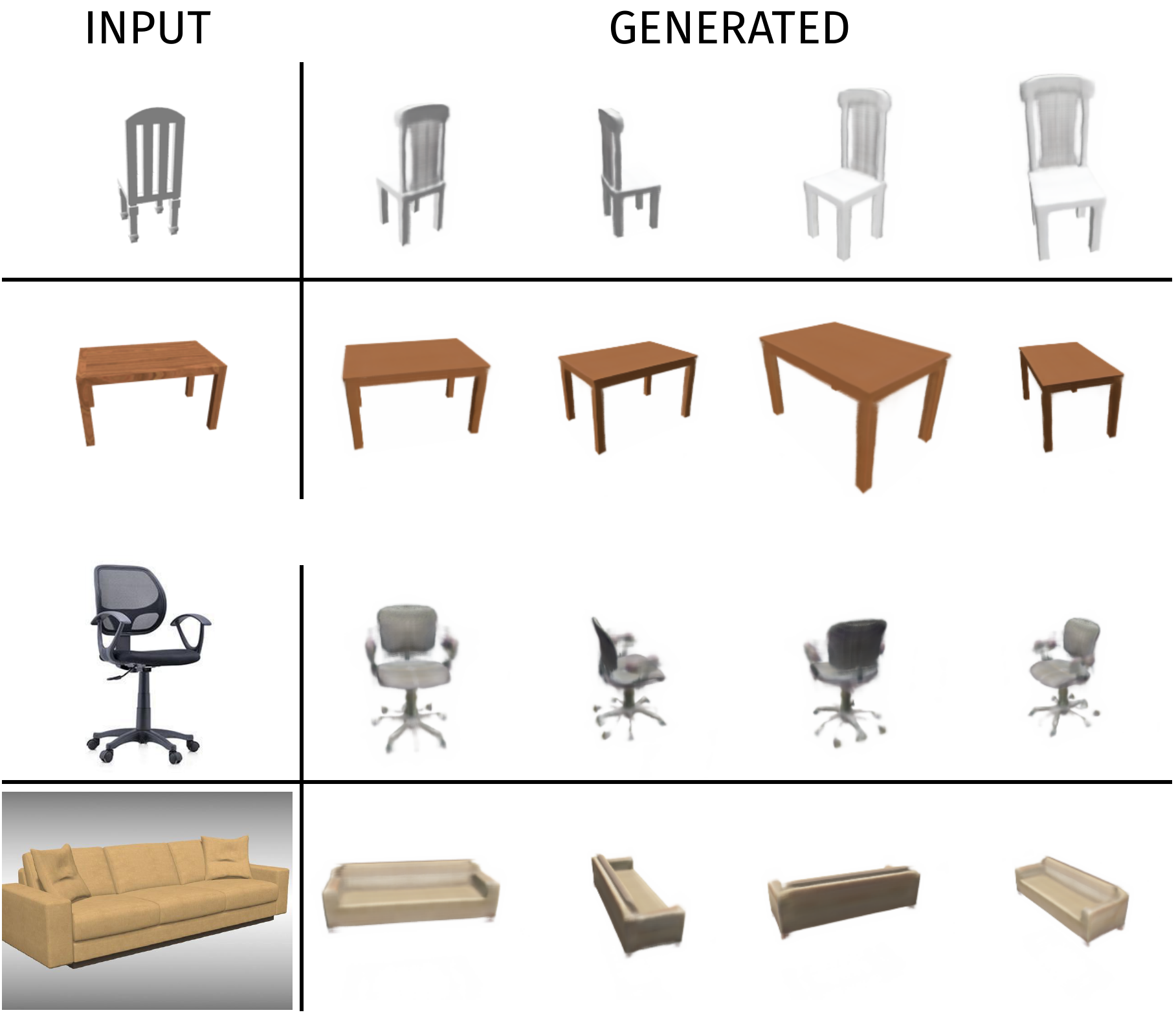}
    \caption{\textbf{Qualitative results of \nerf{} generation from images.} Synthetic images from \shapenet{} \cite{shapenet_render} (top) and real images from DomainNet \cite{domainnet} (bottom).}
    \label{fig:gen-images}
\end{figure}

\begin{figure}[t]
    \centering
    \includegraphics[width=0.9\columnwidth]{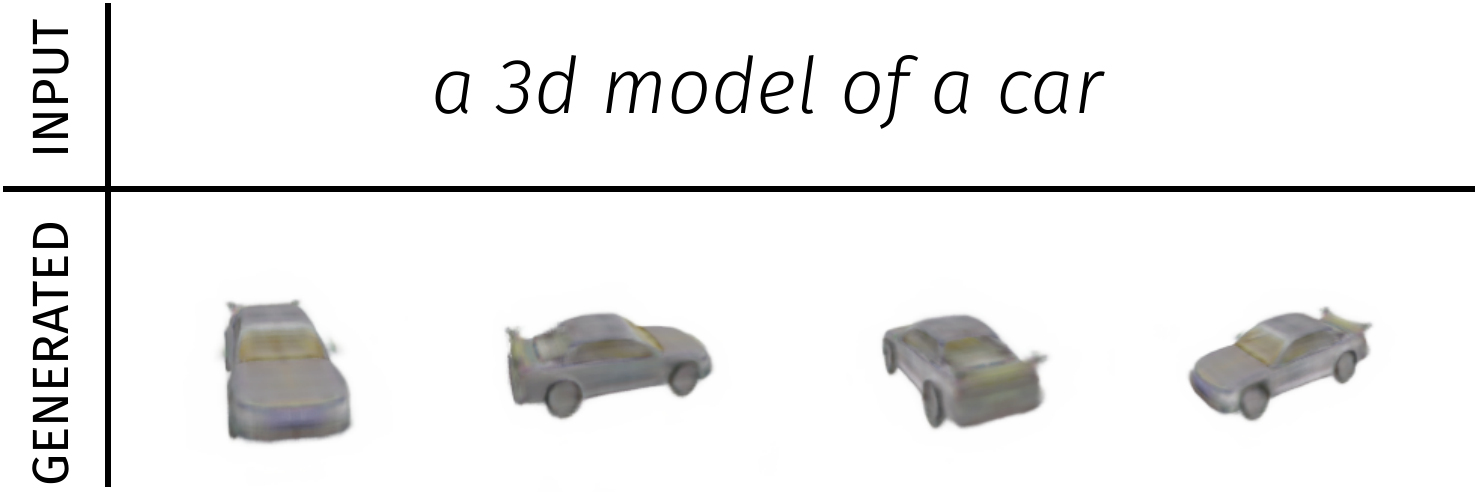}
    \caption{\textbf{Qualitative results of \nerf{} generation from text.}}
    \label{fig:gen-text}
    \vspace{-1mm}
\end{figure}

\paragraph{Generation from images.}
Another application of our approach consists in, given a \nerf{} \cite{nerf} view, \emph{synthesize} new views of the object by leveraging the \nftovec{} \cite{nf2vec} decoder. Specifically, this procedure works as follows: we embed the \nerf{} view with the CLIP \cite{clip} image encoder and give the resulting embedding as input to the trained \cliptonerf{} network, which produces a \nerf{} embedding. The latter can be processed by the \nftovec{} decoder to render arbitrary views of the object. Thus, the embedding plus the decoder can be considered a \nerf{} architecture.
This procedure is illustrated in \cref{fig:gen-method} (top). Qualitative results are shown in \cref{fig:gen-images}, both with images from \shapenet{} \cite{shapenet_render} (top) and real images from DomainNet \cite{domainnet} (bottom).

\paragraph{Generation from text.} Analogously, our framework allows to synthesize new \nerf{} views from text. Given the previous generation pipeline, we replace the CLIP image encoder with the CLIP text encoder (\cref{fig:gen-method} bottom). Qualitative results are shown in \cref{fig:gen-text}.

\section{Limitations and Conclusions}
\label{sec:concl}
Our proposed framework effectively connects \nerf{}, images, and text. We have demonstrated its application in several novel tasks, including \nerf{} retrieval or generation from text and images, as well as zero-shot classification of \nerf s using only network weights.

However, our framework has its limitations. Firstly, the \nftovec{} encoder, trained on \shapenet{}, limits our experiments to \nerf s of synthetic objects only. Additionally, the \nerf{} generation is constrained by the processing capabilities of the \nftovec{} decoder.

In the future, expanding our work to include \nerf s of real objects or scenes would be valuable. 
Learning a shared latent space for \nerf s, images, and text, \eg by jointly training the vision, language, and \nerf{} encoders on larger datasets, could also be a promising direction.
We plan to address these limitations by exploring these ideas in future studies and hope that our framework inspires further advancements in the field.
{
    \small
    \bibliographystyle{ieeenat_fullname}
    \bibliography{main}
}


\end{document}